\pdfoutput=1

\documentclass[11pt]{article}
\usepackage[table]{xcolor}
\usepackage{colortbl}
\usepackage{algpseudocode}
\usepackage{algorithm}
\usepackage[preprint]{acl}
\usepackage{amssymb}

\usepackage{times}
\usepackage{latexsym}
\usepackage{amsmath}
\usepackage{subfigure}
\usepackage{booktabs}
\usepackage{arydshln}
\usepackage{tablefootnote}
\usepackage{multirow}
\usepackage{multicol}

\usepackage[T1]{fontenc}
\usepackage{listings}
\usepackage[utf8]{inputenc}
\usepackage{microtype}
\usepackage{xspace}
\usepackage{inconsolata}
\usepackage{tcolorbox}
\usepackage{graphicx}
\usepackage[normalem]{ulem}
\useunder{\uline}{\ul}{}
\newcommand{\ie}{\emph{i.e.,}\xspace}

\newcommand{\eg}{\emph{e.g.,}\xspace}

\newcommand{\etc}{\textit{etc}\xspace}

\newcommand{\paratitle}[1]{\noindent\textbf{#1}}
\definecolor{cadmiumgreen}{rgb}{0.0, 0.42, 0.24}
\title{ALTER: Augmentation for Large-Table-Based Reasoning}

\author{
    \textbf{Han Zhang \textsuperscript{{1}},
	        Yuheng Ma \textsuperscript{{1}},
                Hanfang Yang \textsuperscript{{1,2}}} \\
	\textsuperscript{1}School of Statistics, Renmin University of China\\
	\textsuperscript{2}Center for Applied Statistics, Renmin University of China\\
    	\texttt{\{hanzhang0816,yma,hyang\}@ruc.edu.cn}\\
}

\begin{document}
\maketitle
\begin{abstract}
While extensive research has explored the use of large language models (LLMs) for table-based reasoning, most approaches struggle with scalability when applied to large tables. To maintain the superior comprehension abilities of LLMs in these scenarios, we introduce \textbf{ALTER} (\underline{A}ugmentation for \underline{L}arge \underline{T}able-bas\underline{E}d \underline{R}easoning)-a framework designed to harness the latent augmentation potential in both free-form natural language (NL) questions, via the query augmentor, and semi-structured tabular data, through the table augmentor. By utilizing only a small subset of relevant data from the table and supplementing it with pre-augmented schema, semantic, and literal information, ALTER achieves outstanding performance on table-based reasoning benchmarks. We also provide a detailed analysis of large-table scenarios, comparing different methods and various partitioning principles. In these scenarios, our method outperforms all other approaches and exhibits robustness and efficiency against perturbations. The code of this paper will be released at {\url{https://github.com/Hanzhang-lang/ALTER}}.

\end{abstract}

\section{Introduction}
Tabular data is one of the fundamental and pivotal semi-structured data types widely used in relational databases, spreadsheets, analysis reports, \etc. Table-based reasoning tasks such as table-based fact verification (FV)~\cite{dataset_feverous, tabfact,ou-liu-2022-learning} and table-based question answering (TQA)~\cite{dataset_hybridqa, wikiTQ,iyyer-etal-2017-search} require sophisticated reasoning over textual, numerical, and logical forms. Besides, the reasoning tasks with large data capacity pose more complexity and challenges to machine intelligence.

Recently, large language models (LLMs) have demonstrated remarkable proficiency in reasoning and inference. The advent of LLMs has spurred a surge in research focusing on their application to tabular data, heralding what can be termed the \textbf{LLM era}~\cite{zhang2024survey, lu2024large}. Despite techniques following the \textbf{pre-LLM era}, such as fine-tuning methods, the latest LLM-based approaches have achieved results that are on par with or surpass those obtained through rule-based or pre-trained language model approaches~\cite{liu-ICLR-22, gu-etal-2022-pasta, jin2022survey}, leveraging the contextual understanding capabilities of LLMs.

Mainstream techniques addressing tabular tasks in the LLM era focus on designing prompts or pipelines that combine multiple instructions with serialized natural language descriptions converted from tables, without requiring additional training. The sequential text data is parsed by the LLMs and transformed into executable code (\eg SQL and Python) using symbolic code generation abilities~\cite{zan-etal-2023-large,cheng2023binding} or direct output for final inference utilizing literal reasoning abilities~\cite{jiang-etal-2023-structgpt,model_tablegpt}.

However, most table-based methods encounter three challenges when analyzing complex large tables. 
\textbf{Firstly}, in the process of converting table cells into natural language descriptions, the entire data is often expected to be included to provide enough comprehensive information~\cite{cheng2023binding}. This approach can sometimes face data leakage issues involving privacy concerns and may fail due to context length limitations. Additionally, the full table content can be long and noisy, leading to unnecessary computational resource consumption. 
\textbf{Secondly}, table reasoning tasks often require numerical reasoning, data preparation, or key cell identification. LLMs alone may lack the robustness to address these tasks directly and can sometimes introduce inaccuracies or hallucinations in their outputs.
As tables grow in size, reasoning about minor or nuanced details becomes even more difficult~\citep{liu2024lost}, and LLMs require careful design to enhance their expandability and robustness in such scenarios. \textbf{Thirdly}, relevant parts needed to derive the answer may be scattered in different places for a complex large-table reasoning task. Therefore, intricate queries cannot be answered in a single glance or with a single execution step using programming languages. Although a couple of methods have optimized for specific issues mentioned above, no approach simultaneously considers all these problems while extending table-based reasoning tasks to large-scale tables.

In consideration of the issues mentioned above, how can we mitigate performance degradation as the size of the table increases? We note that tables are inherently structured; in real-world databases, for instance, tables are well-categorized, and each column feature adheres to certain criteria, including data format, text representation, feature semantics, \etc. Based on these practical observations, in this paper, we propose a framework named \textbf{ALTER} to facilitate the understanding of tables and to scale effectively to large-scale tables. Without utilizing \emph{the entire table data} as contextual information throughout the process, we first generate adaptations about the NL questions with the query augmentor in Section~\ref{sec: query-aug} and interpretations about the table's inherent structure and content with the table augmentor in Section~\ref{sec-aug}. Subsequently, the data is further distilled to filter irrelevant column features. In conjunction with augmented information, the well-organized data is integrated with SQL executors and ultimately transformed into a more accessible format for joint reasoning, adhering to the proposed \emph{augment-filter-execution} procedure.


In summary, our main contributions include: (\romannumeral1) We explore new augmentation methods for queries and tables that are beneficial for table-based reasoning tasks. 
(\romannumeral2) We propose a general framework and a novel \emph{augment-filter-execution} procedure capable of scaling to large tables. (\romannumeral3) We conduct extensive experiments on two table-based reasoning benchmarks and demonstrate the best performance over large-table scenarios.


\section{Related Work}
\paratitle{Large Language Models for Table Reasoning}.
Primary approaches using LLMs to tackle table reasoning tasks involve fine-tuning a foundational model following the pre-LLM era or directly utilizing in-context learning abilities unique to the LLM era. For fine-tuning methods, following the success of mask language modeling (MLM), task-specific fine-tuning methods are designed. For example, TaPas~\cite{herzig-acl-20} extends BERT's~\cite{devlin-etal-2019-bert} architecture and enhances the understanding of tabular data by recovering masked cells. Moreover, models relying on logical codes (\eg SQL) can further enhance the model's reasoning ability. For example, Tapex~\cite{liu-ICLR-22} and OmniTab~\cite{jiang-etal-2022-omnitab} focus on generating SQL queries that are then executed to fetch relevant information.

Prompting technologies such as few-shot learning~\cite{brown2020language}, chain-of-thought reasoning~(COT)~\cite{wei2022chain}, and agent-based methods~\cite{wang2024survey} can be correspondingly applied in table reasoning tasks. \citet{chen2023large} first explores and demonstrates the feasibility of using LLMs in generic reasoning tasks. Binder~\cite{cheng2023binding} shows symbolic languages are also beneficial for complex analysis with prompt methods. Chain-of-Table~\cite{wang2024chain}, inspired by CoT prompting methods, uses tabular data in the reasoning chain as a proxy for intermediate thoughts. ReAcTable~\cite{zhang2023reactable} employs LLMs extending the ReAct framework to reason step-by-step and iteratively generates sub-tables using code executors. Dater~\cite{dater} and DIN-SQL~\cite{din-sql} break down table reasoning into multi-step inference by handcrafting pipeline.

\paratitle{Query Augmentation}. 
In question-answering tasks, query augmentation or query rewrite is a prevalent method to bridge the gap between queries and facts. Within the framework of LLMs, tasks related to Retrieval-Augmented Generation (RAG) often involve various forms of query modification, including query rewriting, disambiguation, and decomposition, \etc~\citep{gao2023retrieval}. RQ-RAG~\citep{chan2024rq} equips the model with multiple capabilities in multi-hop QA tasks. \citet{query-rewrite} proposes \emph{Rewrite-Retrieve-Read} pipeline which adapt the query itself. Step-Back Prompting\cite{zheng2024take} presents a simple technique to derive high-level concepts.

\paratitle{Table Augmentation and Table sampling}. 
Table augmentation involves integrating knowledge and exploring implicit table content. Mainstream methods include incorporating commonsense knowledge~\cite{sui2023tap4llm} from Wikipedia\footnote{\url{https://www.wikipedia.org/}}, obtained through search engines or analytical knowledge~\cite{he2023anameta, model_table_recasting}. \citet{tablemeetllm} relies on LLM itself to augment structural information using internal knowledge. 
\begin{figure*}[t]
\vskip -0.2in
    \centering
    \includegraphics[width=\linewidth]{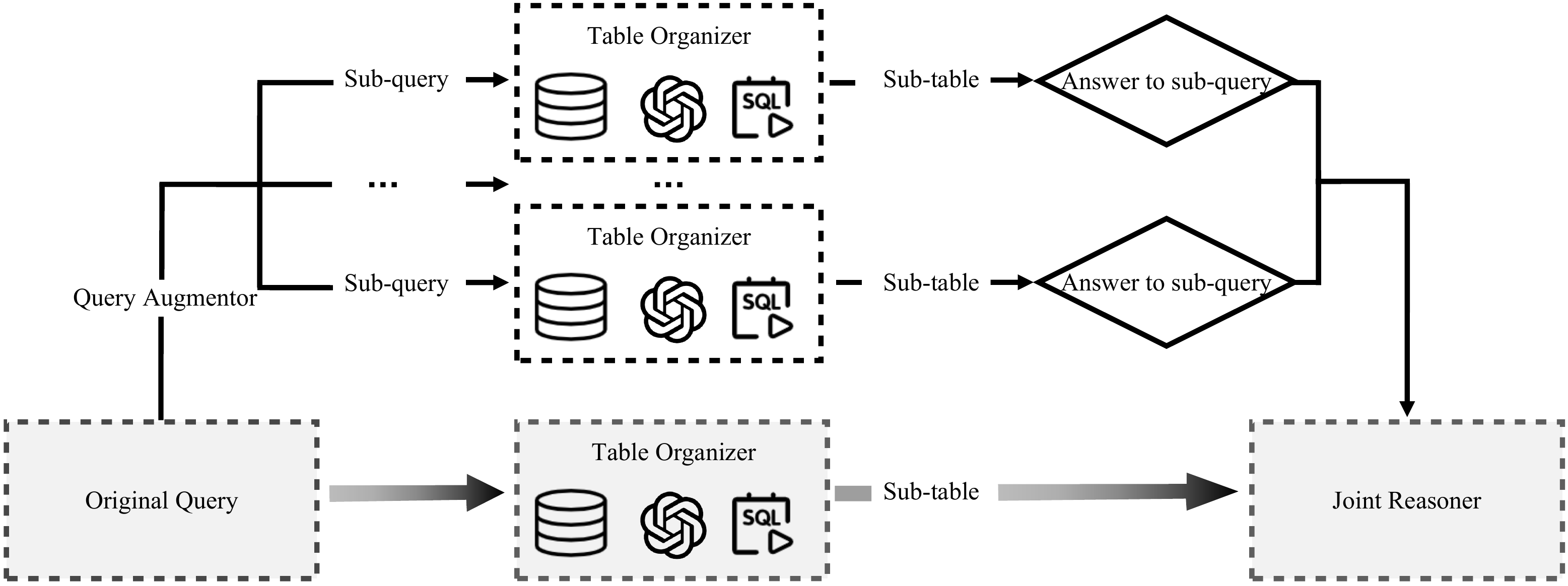}
    \caption{\small{The overview of the ALTER framework for table-based reasoning. The gray background box symbolizes the primary reasoning workflow. Above it, each sub-query generated by the query augmentor is processed in parallel by the table organizer and ultimately transformed into informative demonstrations that aid in understanding the original query. The primary sub-table and relevant information is received by the joint reasoner.}}
    \vskip -0.15in
    \label{fig: alter}
\end{figure*}
\section{Preliminary}
In this section, we introduce the definition of table reasoning tasks. Table reasoning requires reasoning over both free-form natural language(NL) and inherently structured tables. Given the triplet $(T, Q, A)$, where table $T = \{c_i \}_{i=1}^C$,  $C$ represents the number of column features in the table. Note that we do not represent the table in cell format as we expect the table under investigation to adhere to certain norms inherently. $Q$ signifies a query or claim related to the table, and $A$ denotes the answer.

We specifically focus on the table question answering and fact verification tasks. In the table question answering tasks, $Q$ and $A$ correspond to the query and expected answers in natural language form, respectively. In the table fact verification task, $Q$ represents a claim about the table, and the final answer $A \in \{0, 1\}$ where $0$ indicates falsity and $1$ indicates truth regarding the input claim.
\section{Methodology}
\subsection{Overview}
In this work, we assume that semi-structured tabular data is rich in latent information beyond its raw data values. This information suggests that data storage adheres to certain common patterns and field semantics, facilitating the model's inference of the overall data distribution from a minimal sample of data. Inspired by knowledge-fusion models for metadata inference~\citep{he2023anameta} and the internal knowledge-retrieving ability of LLMs~\citep{tablemeetllm}, we utilize LLMs to uncover patterns and semantics within tables, which helps to understand and operate data correctly. The whole workflow is illustrated in Algorithm~\ref{alg: alter} in the appendix. In our framework, we do not include the full content of the table in a prompt; only $K$ rows can be observed. Instead, the reasoning effect is ensured through the elaborate augmented information. The framework seamlessly accommodates large-scale tables, as the model is pre-endowed with comprehensive information about the data structure and content before encountering it. 
As illustrated in Figure~\ref{fig: alter}, our proposed system \textbf{ALTER}, consists of three core components:

$\bullet$ \textbf{Query Augmentor}: This component enhances the original query by generating multiple sub-queries, each examining the original query from different perspectives. Compared to the partial original query, this component comprehensively provides more information through the subsequent table organizer.

$\bullet$ \textbf{Table Organizer}: Given the input query, this component utilizes the \emph{augment-filter-execution} procedure. It first enriches the raw data with augmented table content, then filters the data to retain only highly relevant rows and columns, and finally employs an SQL executor to derive a reasonable and accessible sub-table for final inference.

$\bullet$ \textbf{Joint Reasoner}: This component efficiently performs reasoning and aggregation for the query augmentor and the primary workflow.

\subsection{Query Augmentor}
\label{sec: query-aug}
\begin{figure*}[t]
\vskip -0.2in
    \centering
    \includegraphics[width=\linewidth]{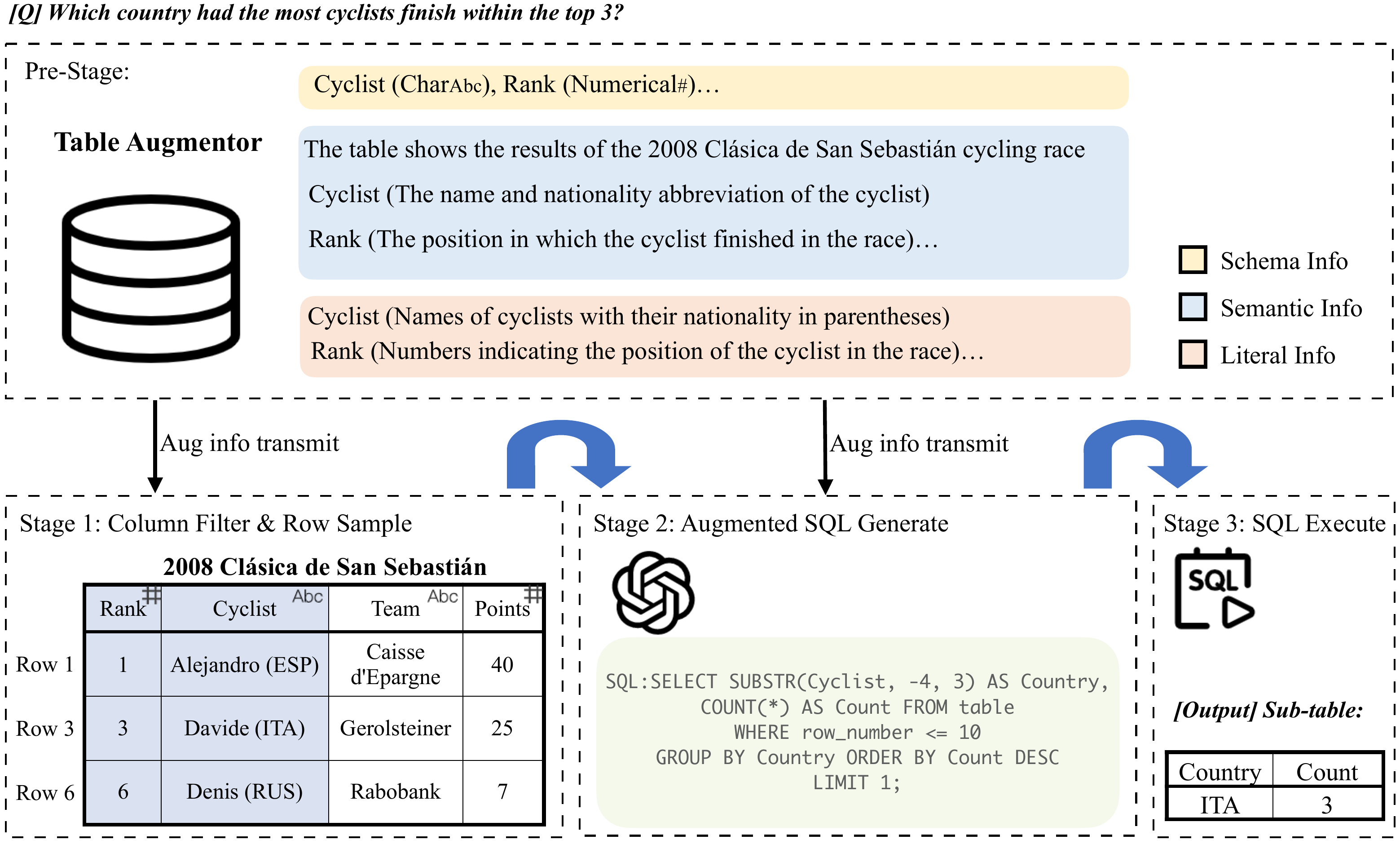}
    \caption{\small{Illustration of the table organizer inside. The augmented information from the table augmentor is utilized in stage 1 and stage 2, enabling the model to correctly locate relevant columns and parse nationalities within the table, ultimately producing the correct execution sub-table.}}
    \label{fig:TableOrganizer}
    \vskip -0.15in
\end{figure*}
\label{query-aug}
One of the primary challenges in naive Question Answering (QA) lies in its direct reliance on the user’s original query as the basis.
Sometimes, the query itself is complex and ambiguous, resulting in subpar effectiveness. In tabular reasoning scenarios, an imprudent query can lead to the model focusing on one partially biased part in the table. We propose a novel improvement method for the query part to mitigate the information loss caused by only access to fractional sub-tables in our downstream process. This approach enables the LLMs to utilize the multi-query technique to attend to different parts within the table through diverse analysis processes. Additionally, based on the results of the sub-queries, it dynamically refines the original query to reduce ambiguity or complexity.

In this work, we propose two query augmentation methods: \emph{step-back augmentation} and \emph{sub-query augmentation}. The step-back prompting method~\cite{zheng2024take} has been empirically validated as effective in the RAG domain. We equip it with sampled sub-table information, which aims to obtain broader and more abstract-level comprehension within the table. LLMs are shown to be stronger at solving sequentially subproblems than directly solving a complex problem~\cite{zhou2022least}. The latter query augmentation method seeks to decompose the information required by the query, enabling LLMs to locate the relevant information more easily in each sub-query.

The reasoning process of all sub-queries in the table organizer is executed in parallel, during which the model utilizes each independent reasoning module to extract information pertinent to answering the original query. Irrelevant information is rejected, and duplicate queries are filtered out.
\subsection{Table Organizer}
The table organizer is the core component of the entire reasoning process. As previously mentioned, throughout the reasoning process, we do not use the entire table data as contextual information. Instead, we further filter the column features of the table, as detailed in Section~\ref{col-filter}, thereby simplifying the transmitted information. To maintain model performance without accessing full data, we employ the \emph{augment-filter-execution} strategy. By pre-analyzing and augmenting the table's schema, semantic, and literal information, sufficient supplementary information required by the query is provided. Given the limited table features, the augmented information does not increase commensurately with the table size. Therefore, our method exhibits strong robustness to variations in table size.

The table organizer primarily encompasses one preparatory stage and three reasoning stages, as illustrated in Figure~\ref{fig:TableOrganizer}. In the preparatory stage, the table augmentor correspondingly mines and enhances information for the downstream process, storing it in advance. In stage 1, based on the schema and semantic information for columns, relevant columns and rows are located. In stage 2, more detailed augmentation information for the filtered columns is considered, including schema, semantic and literal information, \etc. As shown in \citet{cheng2023binding}, high-quality programming language (\eg SQL, Python) can be a powerful tool regarding numerical and logical questions, we rely on SQL as the standard language for querying structured data. Based on the filtered sub-table and incrementally updated augmentation information, executable SQL queries are generated and the final sub-table is retrieved in stage 3.

If the upstream input is a sub-query, the final sub-table will be transformed into an effective response for the sub-query. If the input is the original query, the sub-table, along with the enhancement information from the sub-queries, will be received by the joint reasoner.

\subsubsection{Table Augmentor}
\label{sec-aug}
The table augmentor aims to convey extra information hidden inherently in the table, beyond the raw data itself, to the LLMs. The augmentation process occurs prior to the official reasoning process, as illustrated in Figure~\ref{fig:TableOrganizer}. 

It's worth noting that we can link this process to real large database systems or table applications~\cite{xue2023db}. In standard database systems, extensive work on data cleaning and normalization must be undertaken. During the process, hierarchical augmentation information will be stored and synchronized, including information about the database, tables, and statistical data inside the table. We focus on the latter two levels, as our emphasis lies on the Table QA scenario. In fact, in real-world databases, column names are often stored without semantic meaning and represented by uppercase abbreviations. The data stored may be formatted in equations and abstract symbols, posing challenges in generating SQL queries accurately. Therefore, schema and column feature information need to be predefined and stored in a standardized manner. In this case, we can simplify the process of the table augmentor by migrating augmented information.

In this paper, the extra information table augmentor generates mainly includes the schema information, semantic information, and the literal representation of the table.

Schema information delineates the storage format and interrelationships of database objects. This may include stored procedures at the database level or table relationships like \emph{Foreign Keys} and \emph{Primary Keys} at the table level. For pure data, schema information primarily denotes data types. We extracted three commonly used types in daily analysis: \emph{Numerical, Char, and Date} types. These types are used to standardize data in advance during the process.

The global and feature-specific semantic information enables LLMs to understand the primary content of the table or columns without directly accessing the data. This assists the LLMs in locating the relevant information corresponding to the query and determining the specific domain the table is about. When columns are named using acronyms or aliases, the imparted semantics can be pivotal for analysis.

\citet{wang2024chain} demonstrates SQL queries often fail in accurately parsing the correct format stored in the table and improves it using multiple chain calls. However, the literal representation can explicitly inform the LLMs about the raw data representation format within the table. This facilitates the generation of correctly formatted SQL queries by the LLMs and effectively bridges the gap between complex SQL queries and user questions.
\subsubsection{Column Filter and Row Sample}
\label{col-filter}
Irrelevant table content in the prompt can lead to unnecessary computations and quality regression issues~\cite{sui2023tap4llm}, especially in scenarios involving large tables. We filter column features unrelated to the query and sample relevant rows, which avoids token waste and the introduction of additional bias. Reliance on LLMs to predict the indexes of rows escalates computational costs and budgets~\cite{dater}. Rule-based methods can only match specific patterns, whereas embedding-based methods can leverage semantic and contextual information. Therefore, we initially sample $K$ rows using embedding-based semantic similarity between each row and the utterance, following the practical guide from \citet{sui2023tap4llm}. Subsequently, a powerful LLM is utilized to select columns relevant to the query, excluding irrelevant ones. We utilize the augmented information in Section~\ref{sec-aug} throughout the filtering process.
\subsection{Joint Reasoner}
Given the sub-table derived from the primary workflow (illustrated in Figure~\ref{fig: alter}) and the supplementary information from the query augmentor in \ref{query-aug}, we leverage the step-by-step thought of the LLM to arrive at the final answer.
\section{Experiment}
In this section, we first introduce the datasets and evaluation metrics. We compare ALTER with baseline methods and report the results in Section~\ref{exp: baseline} and Section~\ref{exp: res}. The ablation study and analysis of large-table scenarios are discussed in Section~\ref{exp: ablation} and Section~\ref{exp: large}, respectively. Please refer to Appendix~\ref{app: implementation} for additional implementation details.

\subsection{Datasets and Evaluation Metrics}
\label{exp: datasets}
We evaluate our proposed method on two widely-used table-based reasoning benchmarks, WikiTQ~\citep{wikiTQ} and TabFact~\citep{tabfact}. 

For the table-based fact verification task, we adopt the TabFact dataset, which contains various statements based on Wikipedia tables. We evaluate the dataset using binary classification accuracy on the small-test set containing $1998$ statements with $298$ different tables.

For the table reasoning task,  we adopt WikiTableQuestion (WikiTQ), which contains open-domain tables accompanied by complex questions. We use denotation accuracy as our evaluation metric, which evaluates the predicted answers based on the gold ones. We evaluate our method on the test set containing 4344 samples from 421 different tables.
\subsection{Baselines}
\label{exp: baseline}
\begin{table}[]
\caption{\small{Results of different methods on WikiTQ and TabFact.\protect\footnotemark[1] (We use \underline{underline} to denote the second-best performance, \textbf{bold} to denote the best performance for each region: Pre-LLM era, LLM era with result ensemble and without ensemble)}}
\centering
\vskip -0.1in
\resizebox{1\linewidth}{!}{
\renewcommand{\arraystretch}{1.1}
\setlength{\tabcolsep}{3pt}
\begin{tabular}{lcc}
\toprule
\multicolumn{1}{c}{\multirow{2}{*}{Method}}  & \multicolumn{2}{c}{Acc (\%)}          \\
\multicolumn{1}{c}{}                         & \textsc{WikiTQ} & \textsc{TabFact} \\ \midrule
\rowcolor[rgb]{0.9,0.9,0.9}
\multicolumn{3}{c}{\textit{$\heartsuit$ Pre-LLM era}}                                             \\
TAPEX \cite{liu-ICLR-22}                    & 57.2               & 85.9             \\
TaCube \cite{zhou2022tacube}                 & 60.8               & -                \\
ReasTAP \cite{zhao2022reastap}               & 58.6               & \underline{86.2}             \\
OmniTab \cite{jiang-etal-2022-omnitab}        &\underline{62.7}               & -                \\
CABINET \cite{patnaik2023cabinet}            & \textbf{69.1}               &  -                \\
PASTA \cite{gu-etal-2022-pasta}                     & -                  & \textbf{90.8}             \\ \midrule
\rowcolor[rgb]{0.9,0.9,0.9}
\multicolumn{3}{c}{\textit{$\spadesuit$  LLM era}}                                                 \\
Binder \cite{cheng2023binding}               & 55.1               & 85.1             \\
Dater w SC \cite{dater}                      & 69.0               & 85.4             \\
ReAcTable w s-vote \cite{zhang2023reactable} & 68.0               & 86.1             \\
Mix SC w SC \cite{liu2023rethinking}          & \textbf{73.7}      & -             \\
Chain-of-Table \cite{wang2024chain}          & 67.3               & 86.6             \\
\textbf{ALTER (ours) w SC}                          & \underline{70.4}   & \textbf{87.2}             \\ 
\hdashline
Dater w/o sc \cite{dater}                    & 65.0               & \underline{83.5}                \\
ReAcTable \cite{zhang2023reactable}          & \underline{65.8}   & 83.1             \\
Mix SC w/o SC \cite{liu2023rethinking}         & 64.2               & -                \\
\textbf{ALTER (ours) w/o SC}                        & \textbf{67.4}      &                \textbf{84.3}  \\ \bottomrule
\end{tabular}
}
\label{tab:main res}
\vskip -0.15in
\end{table}

\footnotetext[1]{For the Dater method, we report the results of using the LLM-based method as backbone}

We compare the proposed ALTER with a range of advanced reasoning frameworks for table-based tasks. The baseline methods for comparison can be categorized into two types: mainstream techniques following the pre-LLM era and techniques unique to the LLM era. For the techniques following the pre-LLM era, we select TAPEX~\cite{liu-ICLR-22}, ReasTAP~\cite{zhao2022reastap}, TaCube~\cite{zhou2022tacube}, OmniTab~\cite{jiang-etal-2022-omnitab}, CABINET~\cite{patnaik2023cabinet}. For the techniques unique to the LLM era, we select Binder~\cite{cheng2023binding}, Dater~\cite{dater}, ReAcTable~\cite{zhang2023reactable}, Mix SC~\cite{liu2023rethinking}, Chain-of-Table~\cite{wang2024chain}. Additionally, generating multiple reasoning paths and ultimately choosing the most consistent answer through voting or self-consistency~\cite{wang2022self} can enhance the performance of LLMs. Therefore, for the techniques unique to the LLM era, we report two types of results for those methods employing result ensemble techniques.
\subsection{Results}
\label{exp: res}
We present the results on the WikiTQ and TabFact datasets. The experimental outcomes are summarized in Table~\ref{tab:main res}. From the results, we observe that our ALTER method achieves comparatively outstanding outcomes. Specifically, on the WikiTQ dataset, while the Mix SC method do marginally outperforms our results by aggregating multiple reasoning paths (with 10 sampling times), ALTER still managed to exceed the performance of all other methods under comparison. Notably, ALTER demonstrates the best performance in single-round reasoning among all other methods that utilize result ensemble techniques in the LLM era. This demonstrates the robust performance of our method in reasoning tasks, which can be attributed to the reinforced information provided by the query augmentor and our innovative modular procedure within the table organizer.
\subsection{Ablation Study}
\label{exp: ablation}
We carry out an ablation study to assess the impact of various components on the performance of our methods, as well as to explore the relationship between the pure table data and the inherent augmentation information.

\paratitle{Analysis of the Query Augmentor}. 
\begin{table*}[]
\small
\centering
\vskip -0.15in
\caption{Ablation results of query augmentor on the test sets of WikiTQ and TabFact.}
\resizebox{1\linewidth}{!}{
\renewcommand{\arraystretch}{0.9}
\setlength{\tabcolsep}{7pt}
\begin{tabular}{lcccccc}
\toprule
                \multicolumn{1}{c}{\multirow{2}{*}{\textbf{Methods}}}       & \multicolumn{3}{c}{\textbf{\textsc{TabFact}}}                                        & \multicolumn{3}{c}{\textbf{\textsc{WikiTQ}}}                                      \\
                       & All                     & Simple                 & Hard                    & All                     & Simple                  & Hard                    \\ \midrule
\textbf{ALTER}                  & \textbf{84.3}                    & \textbf{90.7}                    & \textbf{78.2}                   & \textbf{67.4}                  & \textbf{71.2}                    & \textbf{63.4}                    \\
\;\; w/o \textit{step-back} & 82.3 ($\downarrow$ 2.0) & 89.5 ($\downarrow$ 0.9) & 75.4 ($\downarrow$ 2.8)& 64.5 ($\downarrow$ 2.9) & 68.2 ($\downarrow$ 3.0)  & 60.5 ($\downarrow$ 2.9)  \\
\;\; w/o \textit{sub-query} & 82.4 ($\downarrow$ 1.9) & 90.6 ($\downarrow$ 0.1) & 74.6 ($\downarrow$ 3.6)  & 65.4 ($\downarrow$ 2.0)  & 69.7 ($\downarrow$ 1.5)  & 60.8 ($\downarrow$ 2.6) \\ \bottomrule
\end{tabular}
}
\vskip -0.1in
\label{tab: query-abl}
\end{table*}
To analyze the impact of two query augmentation methods in the query augmentor. We conducted experiments on the WikiTQ and TabFact datasets by discarding the step-back augmentation module (denoted as w/o step-back) and the sub-query augmentation module (denoted as w/o sub-query). For each dataset, we further categorized the questions based on the difficulty level, following \citet{dater}. This stratification facilitates a more comprehensive evaluation of each module's impact across different types of questions. The ablation test results are reported in Table~\ref{tab: query-abl}. From the results in the table, it is anticipated that employing both augmentation methods simultaneously yields the best performance under all experimental settings. For WikiTQ datasets, the accuracy of ALTER without step-back/sub-query augmentation drops by $2.9\%/2.0\%$, demonstrating the necessity of augmented information from multi-queries. Furthermore, on the TabFact datasets, both augmentation methods have a much larger impact on hard questions than on simple questions. This indicates that the augmented information provided by the query augmentor is particularly effective in dealing with complex questions. 

\paratitle{Analysis of Pure Data \& Augmentation}.
\begin{table}[htbp]
\small
\centering
\caption{Ablation results of different $K$ values and with or without augmented information from the table augmentor on the WikiTQ and TabFact. (improvement measured against the data in the bottom-left relative position.)}
\resizebox{0.9\linewidth}{!}{
\renewcommand{\arraystretch}{1}
\setlength{\tabcolsep}{5pt}
\begin{tabular}{ccccc}
\toprule
        & \multicolumn{2}{c}{\textbf{\textsc{WikiTQ}}} & \multicolumn{2}{c}{\textbf{\textsc{Tabfact}}} \\
        \cline{2-5} 
        & w/o           & w            & w/o             & w          \\ \midrule
$K = 0$ & 45.5                & 62.2            & 67.1               & 77.2          \\
$K = 1$ & 59.2                & {\color{magenta}65.0 (+1.7)}            & 80.5               & {\color{magenta}82.4 (+0.1)}        \\
$K = 3$ & {\color{magenta}63.3}                & 67.4           & {\color{magenta}82.3 }              & 84.3          \\ \bottomrule
\end{tabular}
}
\vspace{-3mm}
\label{tab:ablation-k}
\end{table}
In our ALTER experiments, we primarily set $K=3$, meaning the model can only access three rows of data relevant to the question throughout the process. To explore the relationship between pure table data and the augmented information in the table organizer, we conducted ablation experiments varying the value of $K$ and the augmentation process. Results are shown in Table~\ref{tab:ablation-k}. We observe that methods utilizing augmented information exhibit significant performance improvements compared to those without augmented information. We also note that the concurrent absence of augmented information and data provision leads to a catastrophic decline in model performance. Notably, on both datasets, using only one row of data with augmented information achieves comparable performance to using three rows of data. Similar trends can also be observed in other settings. This validates that when the model is limited to a small portion of data, the table augmentor serves as a beneficial auxiliary tool, providing additional insights into the table's content.
\subsection{Large Table Analysis}
\label{exp: large}
LLMs often struggle to interpret tables within large-scale scenarios, leading to hallucinations and errors. To the best of our knowledge, nearly all methods encounter a decline in model performance as the table size increases when handling large tables. To demonstrate the effectiveness of the ALTER framework in large-scale scenarios, we compare the performance of our framework across different table sizes in this section. We selected various table partitioning principles and different types of methods for a systematic evaluation. For table partitioning, we employed two approaches based on the token count and the number of cells. For the models, representative methods from both the LLM era and the pre-LLM era are chosen.
\begin{figure}[htbp]
    \centering\includegraphics[width=\linewidth]{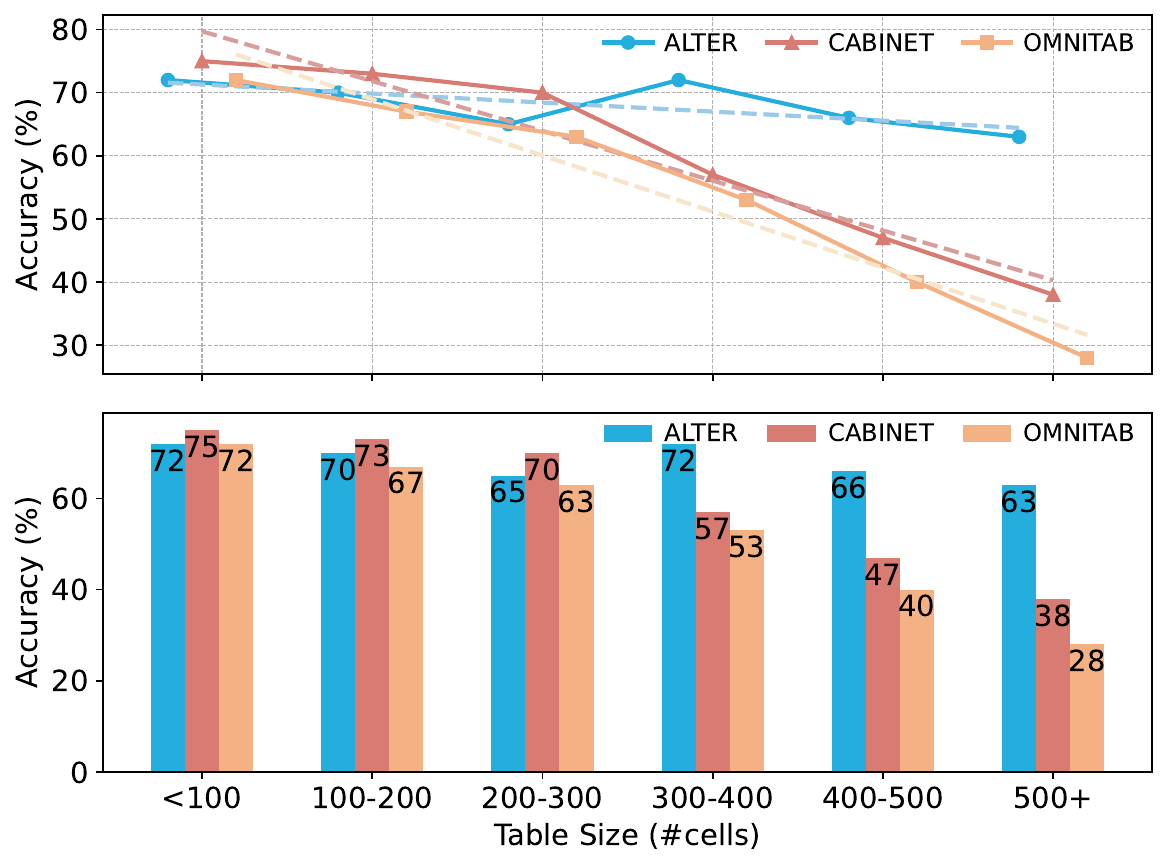}
    \caption{\small{Comparison of methods following pre-LLM era with tables divided by cell count on WikiTQ. In the subplot above, the regression curves of different models are represented by dashed lines in different colors. The regression curve for ALTER exhibits a significantly slower decline rate.}}
    \label{fig:large-table-cell}
\end{figure}

Figure~\ref{fig:large-table-cell} shows the comparison results of ALTER and methods following the pre-LLM era, including CABINET and OMNITAB, partitioning tables in the WikiTQ dataset by the number of cells. In Table~\ref{tab:large-table-token}, we present the results based on different table sizes divided by the token count in the WikiTQ dataset, comparing our method with Dater, Chain-of-TABLE, and Binder unique to the LLM era. Table~\ref{tab:large-table-token} shows that ALTER significantly outperforms all three methods in the LLM era across different table sizes. The performance improvement is particularly noteworthy when dealing with large tables. In Figure~\ref{fig:large-table-cell}, our model demonstrates a much slower performance decline as the model size increases compared to the other two methods. As the size of the table increases, both CABINET and OMNITAB exhibit a monotonous decline in performance. However, our method shows a brief reversal with an increase in performance observed in the intermediate range, indicating the robustness and insensitivity of our approach to changes in table size. Our model significantly outperforms the other two methods when the table size exceeds a certain threshold ($>300$ cells). Specifically, in the $300-400$, $400-500$, and $500+$ cell categories, our model exceeds their performance by at least $15\%$, $19\%$, and $25\%$, respectively. From the results, it is evident that our method exhibits exceptional performance in large tables. 
\begin{table}[htbp]
\vskip -0.1in
\caption{\small{Comparison of methods in the LLM era with tables divided by token count on WikiTQ. (\underline{underline} denotes the second-best performance; \textbf{bold} denotes the best performance)}}
\vskip -0.07in
\centering
\resizebox{1\linewidth}{!}{
\renewcommand{\arraystretch}{1.2}
\setlength{\tabcolsep}{3pt}
\begin{tabular}{llll}
\toprule
\multirow{2}{*}{\textbf{Methods}}   & \multicolumn{3}{c}{\textbf{\textsc{Table Size}}}                                 \\ \cline{2-4} 
                                    & \textbf{Small (<2k)} & \textbf{Medium (2k\textasciitilde4k)} & \textbf{Large (>4k)} \\ \midrule
Binder      & 56.54                & 26.13                     & 6.41                 \\
Dater                  & 62.50               & 42.34                     & 34.62                \\
Chain-of-Table & {\ul 68.13}          & {\ul 52.25}               & {\ul 44.87}          \\
ALTER                       & \textbf{71.67}       & \textbf{65.20}             & \textbf{65.92}        \\ \bottomrule
\end{tabular}
}
\vspace{-3mm}
\label{tab:large-table-token}
\end{table}

\subsection{Robustness and Efficiency Analysis}
\label{sec: robust}
We examined ALTER's robustness to noise perturbations and token efficiency in large-scale scenarios. By adding random rows based on different perturbation factors, we introduced noise to each table in WikiTQ, details of perturbations can be found in Appendix~\ref{sec: perturb}. From Figure~\ref{fig:large-table-perturb}, we illustrate that as the degree of perturbation increases, the proportion of tokens utilized of the whole table by ALTER decreases. It can be observed that the initial fluctuation has the most significant effect, yet our model still outperforms the compared method ($9.8\%$ ALTER \textit{v.s.} $11.4\%$ CABINET). Concurrently, the decline in the framework's performance degree slows down. This indicates that our method efficiently maintains robust performance in large-table scenarios by narrowing down the scope of larger tables.
\begin{figure}[htbp]
    \centering\includegraphics[width=\linewidth]{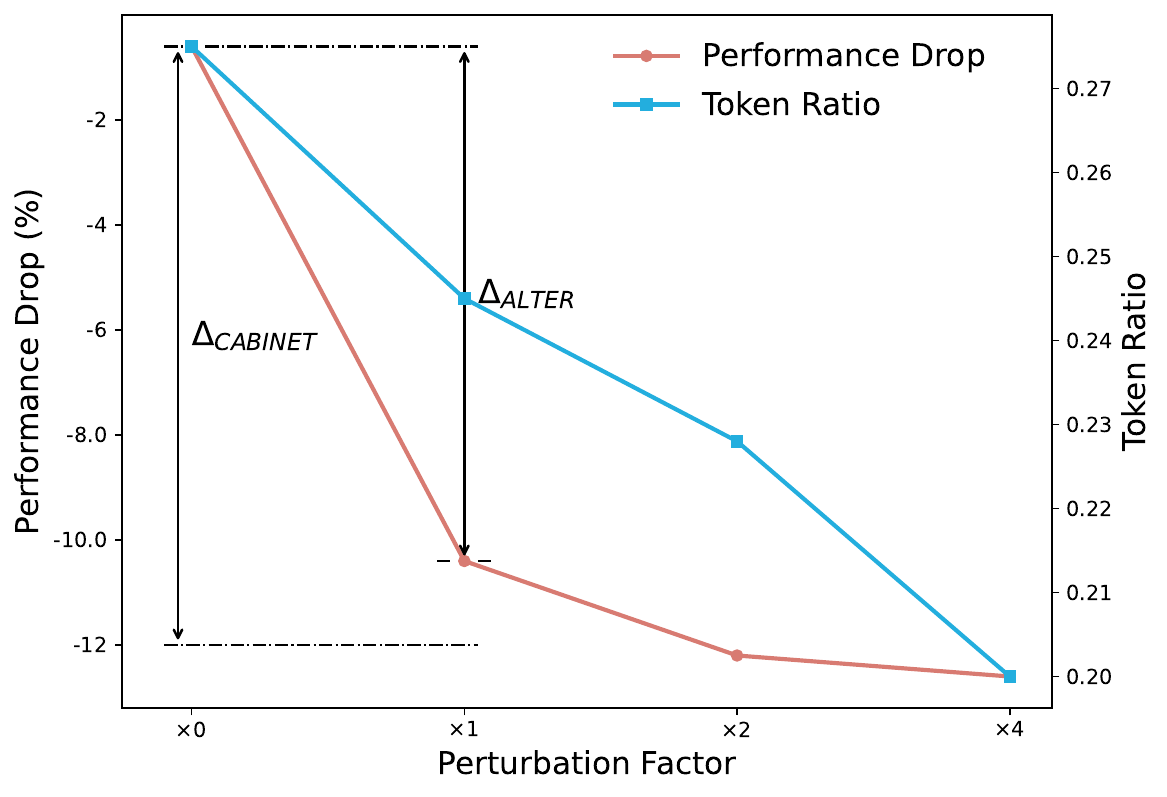}
    \vskip -0.1in
    \caption{\small{Relative performance drop and the ratio of table tokens utilized by ALTER to the total table tokens on WikiTQ as the number of rows added increases by multiples (\ie perturbation factor). The drop for CABINET and ALTER is specifically marked at the factor of $1$.}}
    \label{fig:large-table-perturb}
    \vskip -0.2in
\end{figure}
\subsection{Case Study}
In Appendix~\ref{sec:appendix-case-study}, we present a case study to elucidate the scenarios in which each component of enhanced information within ALTER can facilitate a more profound comprehension of table contents. When addressing complex problems, without the assistance of the augmentation process, the model may focus on biased information or experience hallucinations when generating SQL. However, when the augmented information is explicitly provided, the model can identify the region containing the correct information or generate syntactically correct SQL, thereby delivering accurate responses.

\section{Conclusion}
We propose a framework, namely ALTER, which significantly optimizes model performance on large-scale tables. Within this framework, we extract inherent information pertinent to the questions and tables. By leveraging an \emph{augment-filter-execution} process as the core reasoning workflow, ALTER demonstrates superior performance in handling large tables. We believe ALTER can bridge the gap between table reasoning methodologies and real-world analysis and bring insights into understanding the way LLMs comprehend tables.
\section*{Limitations}
ALTER is designed to generalize to large table reasoning tasks, but our method still faces some limitations. Our approach relies partly on the degree of structured and standardized storage of tables, meaning that if the table structure is totally disordered or lacks a certain level of standardization, our model's performance will degrade, for instance, when headers and data are intermixed. Additionally, the combination methods of different augmented information can be explored further. Due to the page limits, we will leave these explorations for future work.
\bibliography{custom}

\begin{thebibliography}{44}
\providecommand{\natexlab}[1]{#1}

\bibitem[{Aly et~al.(2021)Aly, Guo, Schlichtkrull, Thorne, Vlachos, Christodoulopoulos, Cocarascu, and Mittal}]{dataset_feverous}
Rami Aly, Zhijiang Guo, Michael Schlichtkrull, James Thorne, Andreas Vlachos, Christos Christodoulopoulos, Oana Cocarascu, and Arpit Mittal. 2021.
\newblock Feverous: Fact extraction and verification over unstructured and structured information.
\newblock \emph{arXiv preprint arXiv:2106.05707}.

\bibitem[{Brown et~al.(2020)Brown, Mann, Ryder, Subbiah, Kaplan, Dhariwal, Neelakantan, Shyam, Sastry, Askell et~al.}]{brown2020language}
Tom Brown, Benjamin Mann, Nick Ryder, Melanie Subbiah, Jared~D Kaplan, Prafulla Dhariwal, Arvind Neelakantan, Pranav Shyam, Girish Sastry, Amanda Askell, et~al. 2020.
\newblock Language models are few-shot learners.
\newblock \emph{Advances in neural information processing systems}, 33:1877--1901.

\bibitem[{Chan et~al.(2024)Chan, Xu, Yuan, Luo, Xue, Guo, and Fu}]{chan2024rq}
Chi-Min Chan, Chunpu Xu, Ruibin Yuan, Hongyin Luo, Wei Xue, Yike Guo, and Jie Fu. 2024.
\newblock {RQ-RAG}: Learning to refine queries for retrieval augmented generation.
\newblock \emph{arXiv preprint arXiv:2404.00610}.

\bibitem[{Chen(2023)}]{chen2023large}
Wenhu Chen. 2023.
\newblock \href {https://doi.org/10.18653/v1/2023.findings-eacl.83} {Large language models are few(1)-shot table reasoners}.
\newblock In \emph{Findings of the Association for Computational Linguistics: EACL 2023}, pages 1120--1130, Dubrovnik, Croatia. Association for Computational Linguistics.

\bibitem[{Chen et~al.(2020{\natexlab{a}})Chen, Wang, Chen, Zhang, Wang, Li, Zhou, and Wang}]{tabfact}
Wenhu Chen, Hongmin Wang, Jianshu Chen, Yunkai Zhang, Hong Wang, Shiyang Li, Xiyou Zhou, and William~Yang Wang. 2020{\natexlab{a}}.
\newblock Tabfact: {A} large-scale dataset for table-based fact verification.
\newblock In \emph{8th International Conference on Learning Representations, {ICLR} 2020, Addis Ababa, Ethiopia, April 26-30, 2020}.

\bibitem[{Chen et~al.(2020{\natexlab{b}})Chen, Zha, Chen, Xiong, Wang, and Wang}]{dataset_hybridqa}
Wenhu Chen, Hanwen Zha, Zhiyu Chen, Wenhan Xiong, Hong Wang, and William~Yang Wang. 2020{\natexlab{b}}.
\newblock \href {https://doi.org/10.18653/v1/2020.findings-emnlp.91} {{HybridQA}: {A} {Dataset} of {Multi}-{Hop} {Question} {Answering} over {Tabular} and {Textual} {Data}}.
\newblock In \emph{Findings of the {Association} for {Computational} {Linguistics}: {EMNLP} 2020}, pages 1026--1036, Online. Association for Computational Linguistics.

\bibitem[{Cheng et~al.(2023)Cheng, Xie, Shi, Li, Nadkarni, Hu, Xiong, Radev, Ostendorf, Zettlemoyer, Smith, and Yu}]{cheng2023binding}
Zhoujun Cheng, Tianbao Xie, Peng Shi, Chengzu Li, Rahul Nadkarni, Yushi Hu, Caiming Xiong, Dragomir Radev, Mari Ostendorf, Luke Zettlemoyer, Noah~A. Smith, and Tao Yu. 2023.
\newblock \href {https://openreview.net/forum?id=lH1PV42cbF} {Binding language models in symbolic languages}.
\newblock In \emph{The Eleventh International Conference on Learning Representations, {ICLR} 2023,Kigali, Rwanda, May 1-5, 2023}.

\bibitem[{Devlin et~al.(2019)Devlin, Chang, Lee, and Toutanova}]{devlin-etal-2019-bert}
Jacob Devlin, Ming-Wei Chang, Kenton Lee, and Kristina Toutanova. 2019.
\newblock \href {https://doi.org/10.18653/v1/N19-1423} {{BERT}: Pre-training of deep bidirectional transformers for language understanding}.
\newblock In \emph{Proceedings of the 2019 Conference of the North {A}merican Chapter of the Association for Computational Linguistics: Human Language Technologies, Volume 1 (Long and Short Papers)}, pages 4171--4186, Minneapolis, Minnesota. Association for Computational Linguistics.

\bibitem[{Gao et~al.(2023)Gao, Xiong, Gao, Jia, Pan, Bi, Dai, Sun, and Wang}]{gao2023retrieval}
Yunfan Gao, Yun Xiong, Xinyu Gao, Kangxiang Jia, Jinliu Pan, Yuxi Bi, Yi~Dai, Jiawei Sun, and Haofen Wang. 2023.
\newblock Retrieval-augmented generation for large language models: A survey.
\newblock \emph{arXiv preprint arXiv:2312.10997}.

\bibitem[{Gong et~al.(2020)Gong, Sun, Feng, Qin, Bi, Liu, and Liu}]{model_tablegpt}
Heng Gong, Yawei Sun, Xiaocheng Feng, Bing Qin, Wei Bi, Xiaojiang Liu, and Ting Liu. 2020.
\newblock \href {https://doi.org/10.18653/v1/2020.coling-main.179} {Tablegpt: Few-shot table-to-text generation with table structure reconstruction and content matching}.
\newblock In \emph{Proceedings of the 28th International Conference on Computational Linguistics}, pages 1978--1988, Barcelona, Spain (Online). International Committee on Computational Linguistics.

\bibitem[{Gu et~al.(2022)Gu, Fan, Tang, Nakov, Zhao, and Du}]{gu-etal-2022-pasta}
Zihui Gu, Ju~Fan, Nan Tang, Preslav Nakov, Xiaoman Zhao, and Xiaoyong Du. 2022.
\newblock \href {https://doi.org/10.18653/v1/2022.emnlp-main.331} {{PASTA}: Table-operations aware fact verification via sentence-table cloze pre-training}.
\newblock In \emph{Proceedings of the 2022 Conference on Empirical Methods in Natural Language Processing}, pages 4971--4983, Abu Dhabi, United Arab Emirates. Association for Computational Linguistics.

\bibitem[{He et~al.(2023)He, Zhou, Zhou, Xu, Lv, Li, Shao, Han, Yuan, and Zhang}]{he2023anameta}
Xinyi He, Mengyu Zhou, Mingjie Zhou, Jialiang Xu, Xiao Lv, Tianle Li, Yijia Shao, Shi Han, Zejian Yuan, and Dongmei Zhang. 2023.
\newblock Anameta: A table understanding dataset of field metadata knowledge shared by multi-dimensional data analysis tasks.
\newblock In \emph{Findings of the Association for Computational Linguistics: ACL 2023}, pages 9471--9492.

\bibitem[{Herzig et~al.(2020)Herzig, Nowak, M{\"{u}}ller, Piccinno, and Eisenschlos}]{herzig-acl-20}
Jonathan Herzig, Pawel~Krzysztof Nowak, Thomas M{\"{u}}ller, Francesco Piccinno, and Julian~Martin Eisenschlos. 2020.
\newblock Tapas: Weakly supervised table parsing via pre-training.
\newblock In \emph{Proceedings of the 58th Annual Meeting of the Association for Computational Linguistics, {ACL} 2020, Online, July 5-10, 2020}, pages 4320--4333.

\bibitem[{Iyyer et~al.(2017)Iyyer, Yih, and Chang}]{iyyer-etal-2017-search}
Mohit Iyyer, Wen-tau Yih, and Ming-Wei Chang. 2017.
\newblock \href {https://doi.org/10.18653/v1/P17-1167} {Search-based neural structured learning for sequential question answering}.
\newblock In \emph{Proceedings of the 55th Annual Meeting of the Association for Computational Linguistics (Volume 1: Long Papers)}, pages 1821--1831, Vancouver, Canada. Association for Computational Linguistics.

\bibitem[{Jena et~al.(2022)Jena, Gupta, Shrivastava, and Eisenschlos}]{model_table_recasting}
Aashna Jena, Vivek Gupta, Manish Shrivastava, and Julian Eisenschlos. 2022.
\newblock \href {https://doi.org/10.18653/v1/2022.findings-emnlp.328} {Leveraging data recasting to enhance tabular reasoning}.
\newblock In \emph{Findings of the Association for Computational Linguistics: EMNLP 2022}, pages 4483--4496, Abu Dhabi, United Arab Emirates. Association for Computational Linguistics.

\bibitem[{Jiang et~al.(2023)Jiang, Zhou, Dong, Ye, Zhao, and Wen}]{jiang-etal-2023-structgpt}
Jinhao Jiang, Kun Zhou, Zican Dong, Keming Ye, Xin Zhao, and Ji-Rong Wen. 2023.
\newblock \href {https://doi.org/10.18653/v1/2023.emnlp-main.574} {{S}truct{GPT}: A general framework for large language model to reason over structured data}.
\newblock In \emph{Proceedings of the 2023 Conference on Empirical Methods in Natural Language Processing}, pages 9237--9251, Singapore. Association for Computational Linguistics.

\bibitem[{Jiang et~al.(2022)Jiang, Mao, He, Neubig, and Chen}]{jiang-etal-2022-omnitab}
Zhengbao Jiang, Yi~Mao, Pengcheng He, Graham Neubig, and Weizhu Chen. 2022.
\newblock \href {https://doi.org/10.18653/v1/2022.naacl-main.68} {{O}mni{T}ab: Pretraining with natural and synthetic data for few-shot table-based question answering}.
\newblock In \emph{Proceedings of the 2022 Conference of the North American Chapter of the Association for Computational Linguistics: Human Language Technologies}, pages 932--942, Seattle, United States. Association for Computational Linguistics.

\bibitem[{Jin et~al.(2022)Jin, Siebert, Li, and Chen}]{jin2022survey}
Nengzheng Jin, Joanna Siebert, Dongfang Li, and Qingcai Chen. 2022.
\newblock A survey on table question answering: recent advances.
\newblock In \emph{China Conference on Knowledge Graph and Semantic Computing}, pages 174--186. Springer.

\bibitem[{Johnson et~al.(2019)Johnson, Douze, and J{\'e}gou}]{johnson2019billion}
Jeff Johnson, Matthijs Douze, and Herv{\'e} J{\'e}gou. 2019.
\newblock Billion-scale similarity search with {GPUs}.
\newblock \emph{IEEE Transactions on Big Data}, 7(3):535--547.

\bibitem[{Liu et~al.(2024)Liu, Lin, Hewitt, Paranjape, Bevilacqua, Petroni, and Liang}]{liu2024lost}
Nelson~F Liu, Kevin Lin, John Hewitt, Ashwin Paranjape, Michele Bevilacqua, Fabio Petroni, and Percy Liang. 2024.
\newblock Lost in the middle: How language models use long contexts.
\newblock \emph{Transactions of the Association for Computational Linguistics}, 12:157--173.

\bibitem[{Liu et~al.(2022)Liu, Chen, Guo, Ziyadi, Lin, Chen, and Lou}]{liu-ICLR-22}
Qian Liu, Bei Chen, Jiaqi Guo, Morteza Ziyadi, Zeqi Lin, Weizhu Chen, and Jian{-}Guang Lou. 2022.
\newblock {TAPEX:} table pre-training via learning a neural {SQL} executor.
\newblock In \emph{The Tenth International Conference on Learning Representations, {ICLR} 2022, Virtual Event, April 25-29, 2022}.

\bibitem[{Liu et~al.(2023)Liu, Wang, and Chen}]{liu2023rethinking}
Tianyang Liu, Fei Wang, and Muhao Chen. 2023.
\newblock Rethinking tabular data understanding with large language models.
\newblock \emph{arXiv preprint arXiv:2312.16702}.

\bibitem[{Lu et~al.(2024)Lu, Zhang, Zhang, and Chen}]{lu2024large}
Weizheng Lu, Jiaming Zhang, Jing Zhang, and Yueguo Chen. 2024.
\newblock Large language model for table processing: A survey.
\newblock \emph{arXiv preprint arXiv:2402.05121}.

\bibitem[{Ma et~al.(2023)Ma, Gong, He, Zhao, and Duan}]{query-rewrite}
Xinbei Ma, Yeyun Gong, Pengcheng He, Hai Zhao, and Nan Duan. 2023.
\newblock \href {https://doi.org/10.18653/v1/2023.emnlp-main.322} {Query rewriting in retrieval-augmented large language models}.
\newblock In \emph{Proceedings of the 2023 Conference on Empirical Methods in Natural Language Processing}, pages 5303--5315, Singapore. Association for Computational Linguistics.

\bibitem[{Ou and Liu(2022)}]{ou-liu-2022-learning}
Suixin Ou and Yongmei Liu. 2022.
\newblock \href {https://doi.org/10.18653/v1/2022.acl-long.525} {Learning to generate programs for table fact verification via structure-aware semantic parsing}.
\newblock In \emph{Proceedings of the 60th Annual Meeting of the Association for Computational Linguistics (Volume 1: Long Papers)}, pages 7624--7638, Dublin, Ireland. Association for Computational Linguistics.

\bibitem[{Pasupat and Liang(2015)}]{wikiTQ}
Panupong Pasupat and Percy Liang. 2015.
\newblock Compositional semantic parsing on semi-structured tables.
\newblock In \emph{Proceedings of the 53rd Annual Meeting of the Association for Computational Linguistics and the 7th International Joint Conference on Natural Language Processing of the Asian Federation of Natural Language Processing, {ACL} 2015, July 26-31, 2015, Beijing, China, Volume 1: Long Papers}, pages 1470--1480.

\bibitem[{Patnaik et~al.(2024)Patnaik, Changwal, Aggarwal, Bhatia, Kumar, and Krishnamurthy}]{patnaik2023cabinet}
Sohan Patnaik, Heril Changwal, Milan Aggarwal, Sumit Bhatia, Yaman Kumar, and Balaji Krishnamurthy. 2024.
\newblock {CABINET}: Content relevance-based noise reduction for table question answering.
\newblock In \emph{The Twelfth International Conference on Learning Representations,{ICLR} 2024, Vienna, Austria, May 7-11, 2024}.

\bibitem[{Pourreza and Rafiei(2023)}]{din-sql}
Mohammadreza Pourreza and Davood Rafiei. 2023.
\newblock \href {https://proceedings.neurips.cc/paper_files/paper/2023/file/72223cc66f63ca1aa59edaec1b3670e6-Paper-Conference.pdf} {{DIN-SQL}: Decomposed in-context learning of text-to-sql with self-correction}.
\newblock In \emph{Advances in Neural Information Processing Systems}, volume~36, pages 36339--36348. Curran Associates, Inc.

\bibitem[{Sui et~al.(2024)Sui, Zhou, Zhou, Han, and Zhang}]{tablemeetllm}
Yuan Sui, Mengyu Zhou, Mingjie Zhou, Shi Han, and Dongmei Zhang. 2024.
\newblock \href {https://doi.org/10.1145/3616855.3635752} {Table meets llm: Can large language models understand structured table data? a benchmark and empirical study}.
\newblock In \emph{Proceedings of the 17th ACM International Conference on Web Search and Data Mining}, WSDM '24, page 645–654, New York, NY, USA. Association for Computing Machinery.

\bibitem[{Sui et~al.(2023)Sui, Zou, Zhou, He, Du, Han, and Zhang}]{sui2023tap4llm}
Yuan Sui, Jiaru Zou, Mengyu Zhou, Xinyi He, Lun Du, Shi Han, and Dongmei Zhang. 2023.
\newblock Tap4llm: Table provider on sampling, augmenting, and packing semi-structured data for large language model reasoning.
\newblock \emph{arXiv preprint arXiv:2312.09039}.

\bibitem[{Wang et~al.(2024{\natexlab{a}})Wang, Ma, Feng, Zhang, Yang, Zhang, Chen, Tang, Chen, Lin et~al.}]{wang2024survey}
Lei Wang, Chen Ma, Xueyang Feng, Zeyu Zhang, Hao Yang, Jingsen Zhang, Zhiyuan Chen, Jiakai Tang, Xu~Chen, Yankai Lin, et~al. 2024{\natexlab{a}}.
\newblock A survey on large language model based autonomous agents.
\newblock \emph{Frontiers of Computer Science}, 18(6):1--26.

\bibitem[{Wang et~al.(2022)Wang, Wei, Schuurmans, Le, Chi, Narang, Chowdhery, and Zhou}]{wang2022self}
Xuezhi Wang, Jason Wei, Dale Schuurmans, Quoc Le, Ed~Chi, Sharan Narang, Aakanksha Chowdhery, and Denny Zhou. 2022.
\newblock Self-consistency improves chain of thought reasoning in language models.
\newblock \emph{arXiv preprint arXiv:2203.11171}.

\bibitem[{Wang et~al.(2024{\natexlab{b}})Wang, Zhang, Li, Eisenschlos, Perot, Wang, Miculicich, Fujii, Shang, Lee, and Pfister}]{wang2024chain}
Zilong Wang, Hao Zhang, Chun-Liang Li, Julian~Martin Eisenschlos, Vincent Perot, Zifeng Wang, Lesly Miculicich, Yasuhisa Fujii, Jingbo Shang, Chen-Yu Lee, and Tomas Pfister. 2024{\natexlab{b}}.
\newblock \href {https://openreview.net/forum?id=4L0xnS4GQM} {Chain-of-table: Evolving tables in the reasoning chain for table understanding}.
\newblock In \emph{The Twelfth International Conference on Learning Representations,{ICLR} 2024,Vienna, Austria, May 7-11, 2024}.

\bibitem[{Wei et~al.(2022)Wei, Wang, Schuurmans, Bosma, Xia, Chi, Le, Zhou et~al.}]{wei2022chain}
Jason Wei, Xuezhi Wang, Dale Schuurmans, Maarten Bosma, Fei Xia, Ed~Chi, Quoc~V Le, Denny Zhou, et~al. 2022.
\newblock Chain-of-thought prompting elicits reasoning in large language models.
\newblock \emph{Advances in neural information processing systems}, 35:24824--24837.

\bibitem[{Xiao et~al.(2023)Xiao, Liu, Zhang, and Muennighoff}]{bge_embedding}
Shitao Xiao, Zheng Liu, Peitian Zhang, and Niklas Muennighoff. 2023.
\newblock \href {https://arxiv.org/abs/2309.07597} {C-pack: Packaged resources to advance general chinese embedding}.
\newblock \emph{Preprint}, arXiv:2309.07597.

\bibitem[{Xue et~al.(2023)Xue, Jiang, Shi, Cheng, Chen, Yang, Zhang, He, Zhang, Wei et~al.}]{xue2023db}
Siqiao Xue, Caigao Jiang, Wenhui Shi, Fangyin Cheng, Keting Chen, Hongjun Yang, Zhiping Zhang, Jianshan He, Hongyang Zhang, Ganglin Wei, et~al. 2023.
\newblock {DB-GPT}: Empowering database interactions with private large language models.
\newblock \emph{arXiv preprint arXiv:2312.17449}.

\bibitem[{Ye et~al.(2023)Ye, Hui, Yang, Li, Huang, and Li}]{dater}
Yunhu Ye, Binyuan Hui, Min Yang, Binhua Li, Fei Huang, and Yongbin Li. 2023.
\newblock \href {https://doi.org/10.1145/3539618.3591708} {Large language models are versatile decomposers: Decomposing evidence and questions for table-based reasoning}.
\newblock In \emph{Proceedings of the 46th International ACM SIGIR Conference on Research and Development in Information Retrieval}, SIGIR '23, page 174–184, New York, NY, USA. Association for Computing Machinery.

\bibitem[{Zan et~al.(2023)Zan, Chen, Zhang, Lu, Wu, Guan, Yongji, and Lou}]{zan-etal-2023-large}
Daoguang Zan, Bei Chen, Fengji Zhang, Dianjie Lu, Bingchao Wu, Bei Guan, Wang Yongji, and Jian-Guang Lou. 2023.
\newblock \href {https://doi.org/10.18653/v1/2023.acl-long.411} {Large language models meet {NL}2{C}ode: A survey}.
\newblock In \emph{Proceedings of the 61st Annual Meeting of the Association for Computational Linguistics (Volume 1: Long Papers)}, pages 7443--7464, Toronto, Canada. Association for Computational Linguistics.

\bibitem[{Zhang et~al.(2024)Zhang, Wang, Dou, Zhu, and Che}]{zhang2024survey}
Xuanliang Zhang, Dingzirui Wang, Longxu Dou, Qingfu Zhu, and Wanxiang Che. 2024.
\newblock A survey of table reasoning with large language models.
\newblock \emph{arXiv preprint arXiv:2402.08259}.

\bibitem[{Zhang et~al.(2023)Zhang, Henkel, Floratou, Cahoon, Deep, and Patel}]{zhang2023reactable}
Yunjia Zhang, Jordan Henkel, Avrilia Floratou, Joyce Cahoon, Shaleen Deep, and Jignesh~M Patel. 2023.
\newblock Reactable: Enhancing react for table question answering.
\newblock \emph{arXiv preprint arXiv:2310.00815}.

\bibitem[{Zhao et~al.(2022)Zhao, Nan, Qi, Zhang, and Radev}]{zhao2022reastap}
Yilun Zhao, Linyong Nan, Zhenting Qi, Rui Zhang, and Dragomir Radev. 2022.
\newblock Reastap: Injecting table reasoning skills during pre-training via synthetic reasoning examples.
\newblock In \emph{2022 Conference on Empirical Methods in Natural Language Processing, EMNLP 2022}.

\bibitem[{Zheng et~al.(2024)Zheng, Mishra, Chen, Cheng, Chi, Le, and Zhou}]{zheng2024take}
Huaixiu~Steven Zheng, Swaroop Mishra, Xinyun Chen, Heng-Tze Cheng, Ed~H. Chi, Quoc~V Le, and Denny Zhou. 2024.
\newblock \href {https://openreview.net/forum?id=3bq3jsvcQ1} {Take a step back: Evoking reasoning via abstraction in large language models}.
\newblock In \emph{The Twelfth International Conference on Learning Representations,{ICLR} 2024, Vienna, Austria, May 7-11, 2024}.

\bibitem[{Zhou et~al.(2022{\natexlab{a}})Zhou, Sch{\"a}rli, Hou, Wei, Scales, Wang, Schuurmans, Cui, Bousquet, Le et~al.}]{zhou2022least}
Denny Zhou, Nathanael Sch{\"a}rli, Le~Hou, Jason Wei, Nathan Scales, Xuezhi Wang, Dale Schuurmans, Claire Cui, Olivier Bousquet, Quoc Le, et~al. 2022{\natexlab{a}}.
\newblock Least-to-most prompting enables complex reasoning in large language models.
\newblock \emph{arXiv preprint arXiv:2205.10625}.

\bibitem[{Zhou et~al.(2022{\natexlab{b}})Zhou, Hu, Dong, Cheng, Cheng, Han, and Zhang}]{zhou2022tacube}
Fan Zhou, Mengkang Hu, Haoyu Dong, Zhoujun Cheng, Fan Cheng, Shi Han, and Dongmei Zhang. 2022{\natexlab{b}}.
\newblock Tacube: Pre-computing data cubes for answering numerical-reasoning questions over tabular data.
\newblock In \emph{Proceedings of the 2022 Conference on Empirical Methods in Natural Language Processing}, pages 2278--2291.

\end{thebibliography}

\appendix
\clearpage

\section{Implementation Details}
\label{app: implementation}
All experiments in this paper were conducted on GPU clusters with 4 NVIDIA A100 GPUs. We employ GPT-3.5-turbo as our large language model backbone in all experiments. To ensure consistent results, we apply a self-consistency technique with $5$ sampling times for each benchmark dataset. For the embedding model in Section \ref{col-filter}, we utilize bge-large-en model~\cite{bge_embedding} and employ FAISS~\cite{johnson2019billion} for efficient similarity search.
\section{Case Study}
\label{sec:appendix-case-study}
\begin{figure}[h]
\vskip -0.1in
    \centering\includegraphics[width=\linewidth]{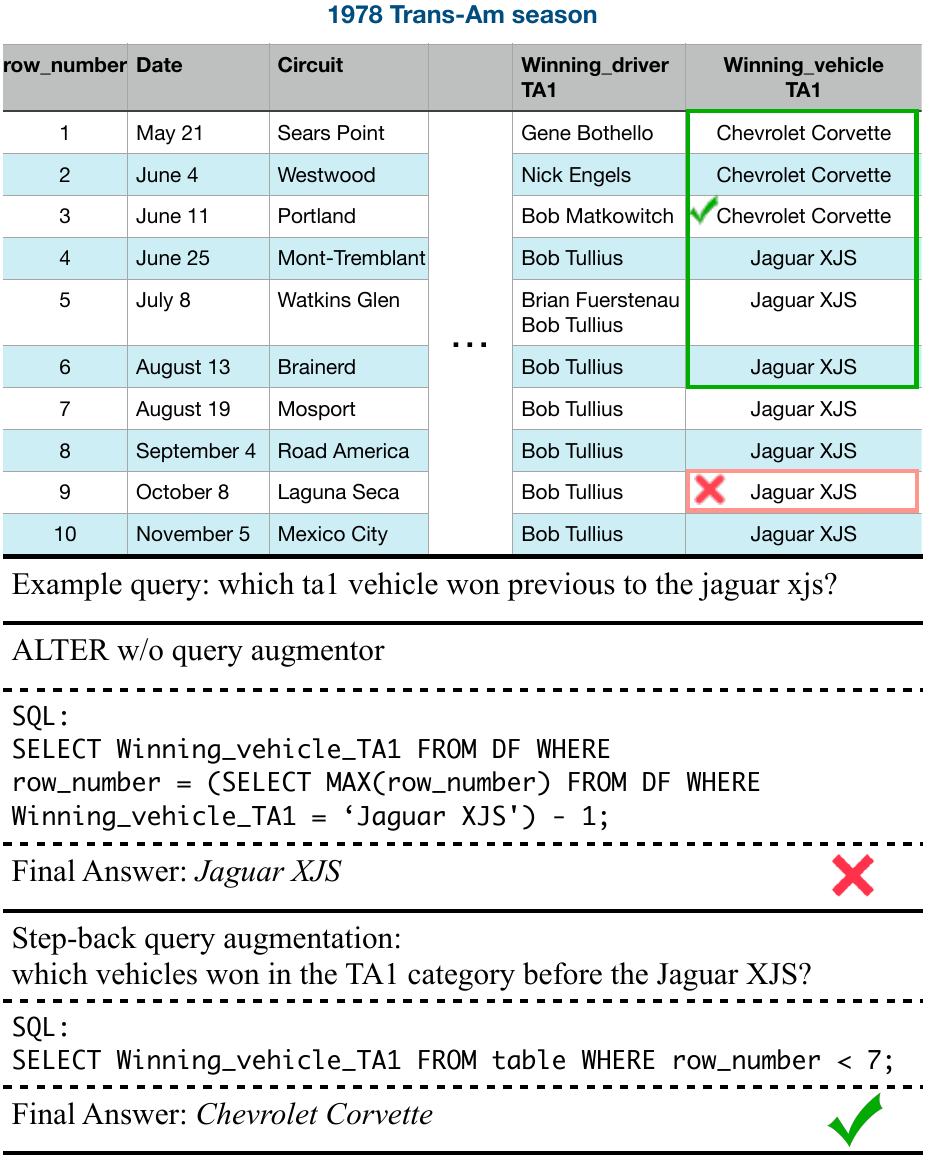}
    \vskip -0.1in
    \caption{Intuitive example for step-back query augmentation, where ALTER correctly answers the query utilizing broader information compared to directly output SQL based on the original query.}
    \label{fig: case-study-step}
\end{figure}

In Figure~\ref{fig: case-study-step}, the input question asks for the \emph{vehicle preceding the Jaguar XJS}. When filtered table data is directly provided, the SQL output for the original query only attends to the second last row of the table. This indicates that the model has observed biased data, incorrectly assuming that the vehicle Jaguar XJS appears only once. However, through step-back query augmentation, the query is reframed, and the model generates a more general SQL query, acquiring more results and thus arriving at the correct answer. This demonstrates that step-back query augmentation enables the model to access a broader scope of information.

\begin{figure}[t]
\vskip 0.98in
    \centering\includegraphics[width=\linewidth]{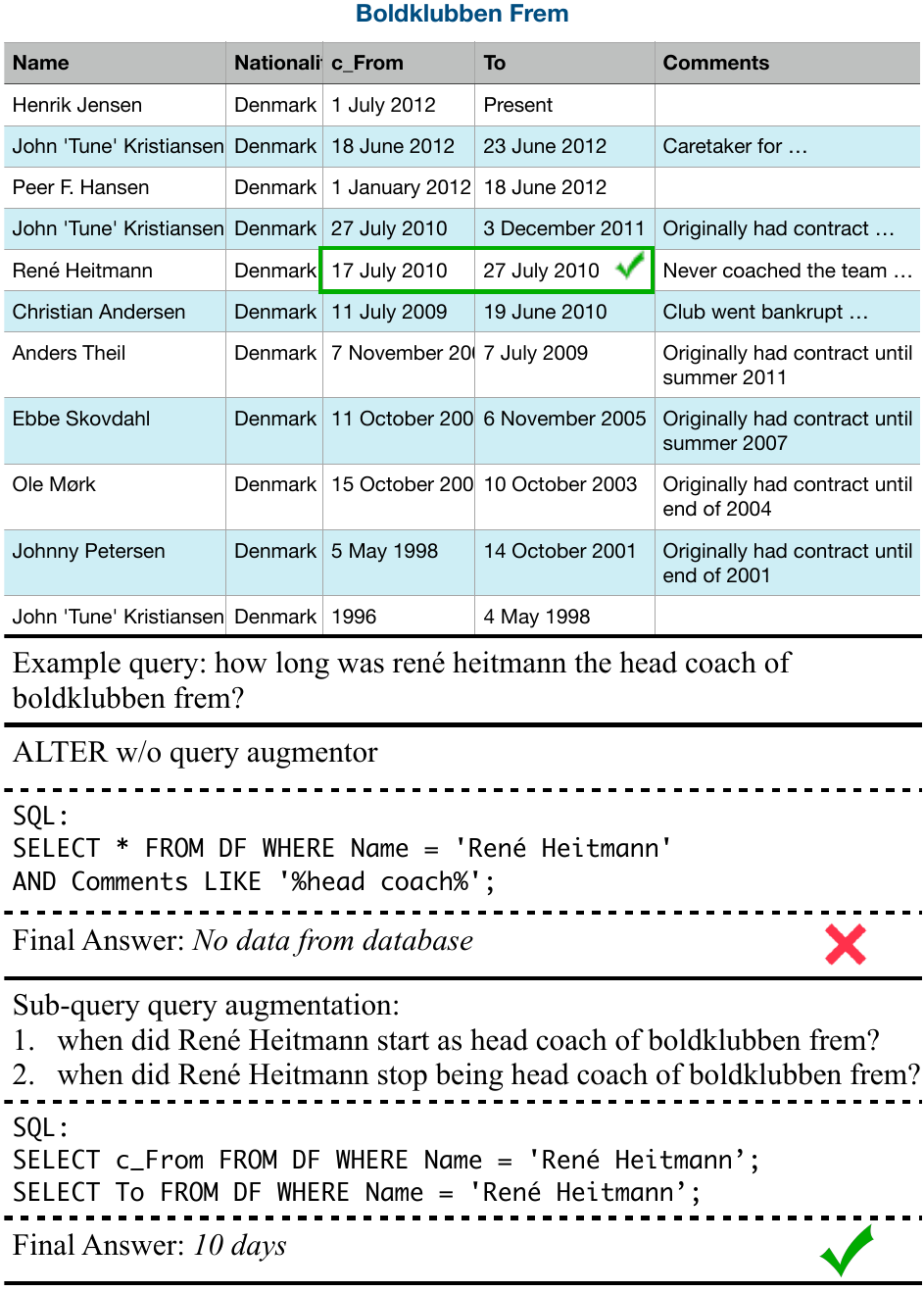}
    \caption{Intuitive example for sub-query query augmentation, where ALTER correctly answers the query utilizing sub-queries compared to directly output SQL based on the original query.}
    \label{fig: case-study-sub}
\end{figure}

In Figure~\ref{fig: case-study-sub}, the input query seeks to determine the tenure of \emph{René Heitmann} as head coach. This involves operations on two distinct feature columns. By decomposing the original query into sub-queries, the difficulty is reduced, allowing the model to accurately retrieve the corresponding information and ultimately compute the correct result.

\begin{figure}[t]
\vskip -2.37in
    \centering\includegraphics[width=\linewidth]{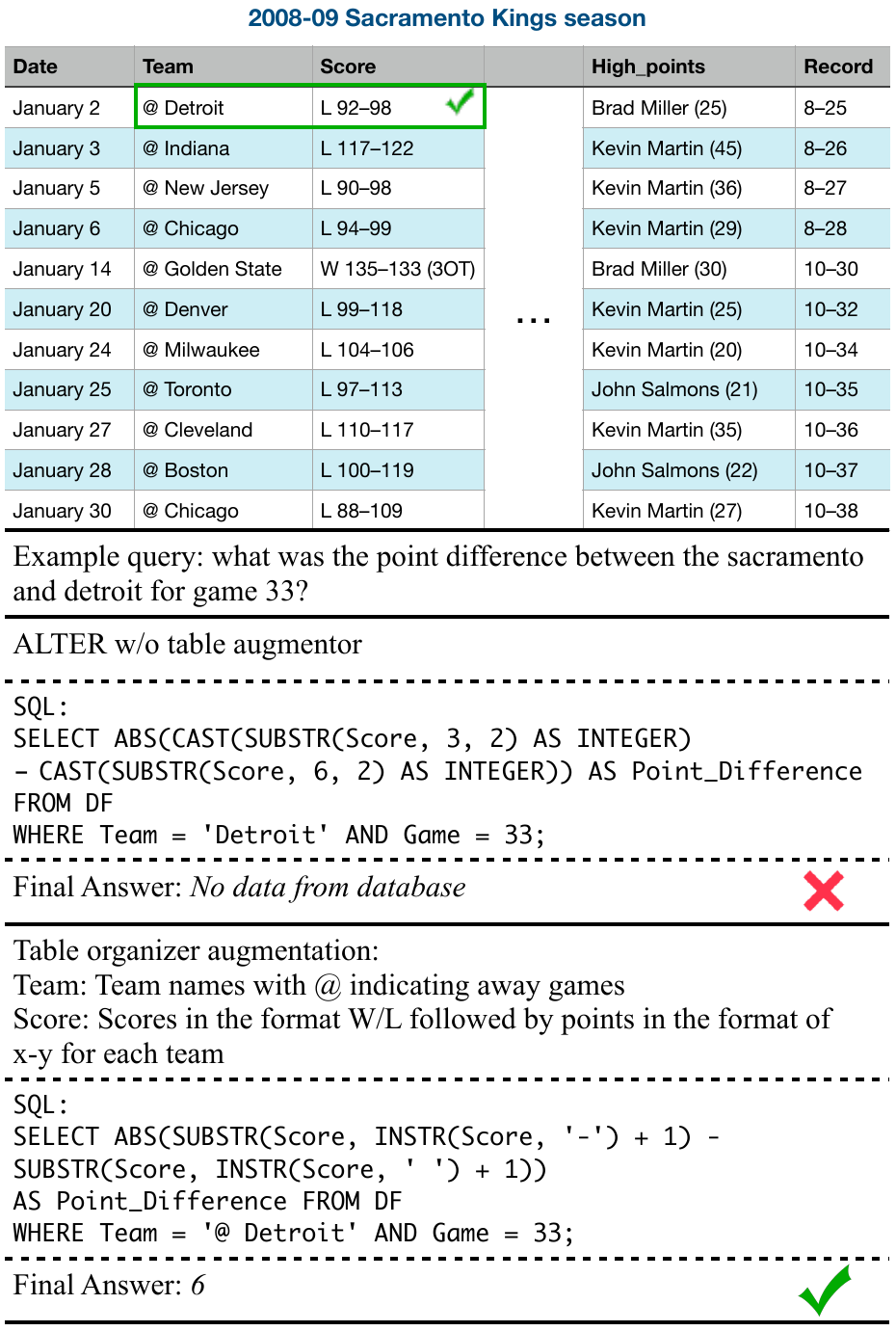}
    \caption{Intuitive example for table augmentor, where ALTER correctly answers the query utilizing information about data format and composition compared to directly output SQL without any augmentation information.}
    \label{fig: case-study-aug}
\end{figure}

In Figure~\ref{fig: case-study-aug}, the input query seeks to determine the score differential for the team \emph{Detroit}. Without relying on the augmented information from the table augmentor, the model fails to correctly capture the name in the \emph{Team} column and cannot accurately extract the score values in the \emph{Score} column. After incorporating the augmented information, the model can generate syntactically correct SQL and extract the needed data.
\section{Details of Table Perturbation}
\label{sec: perturb}
In Section~\ref{sec: robust}, we discussed the robustness and efficiency of ALTER. We provide details of the perturbations implemented. We insert noise into a table by adding rows based on the size of the table, following the row adding steps in~\citet{patnaik2023cabinet}. However, we do not randomly extract values from other tables, as this would compromise the pre-augmented schema standardization. Based on the augmented schema information, we randomly generated data for three types of features: Date, Numerical, and Char. We believe the disturbance intensity is quite similar for the model compared to the previous approach. Based on the number of cells ($\#cells = N$) in the table, the exact scheme of the $n$ rows inserted is as follows:
(\romannumeral1) $n=1 \; \text{if}\; N \leq 150$, (\romannumeral2) $n=2 \; \text{if}\; 150 < N \leq 300$, (\romannumeral3) $n=4 \; \text{if}\; 300 < N \leq 450$, (\romannumeral4) $n=8 \; \text{if}\; N \geq 450$. Additionally, for each of these categories, we vary the degrees of perturbation by multiplying the number of added rows by 1, 2, and 4 times (\ie perturbation factor used in Figure~\ref{fig:large-table-perturb}). 
\section{Prompts}
\label{sec: prompt}
We provide the prompt templates for different augmentation methods used within the ALTER framework. See Figure~\ref{fig:prompt-query} for two query augmentation methods and Figure~\ref{fig:prompt-aug} for different augmentations used in table augmentor. In these templates, the red text serves as a placeholder for specific input. The in-context few-shot examples are selected from the training or validation set for each task. The sub-tables are serialized into HTML format throughout the experiments.

\begin{figure*}[h]
\begin{tcolorbox}[left=1mm,right=1mm,top=0.mm, bottom=0mm,colback=white]
\begin{lstlisting}
=========================== /*** Step-back Augmentor ***/ ===========================
Below is a sub-table with rows randomly sampled from the original table. Based on the sub-table, your task is to step back and paraphrase a question to a more generic step-back question, which is easier to answer.

\{{In-context examples}\}

Sub-table: \{{Sub-table}\}
Query:\{{Query}\}
New query: 
=========================== /*** Sub-query Augmentor ***/ ===========================
You are capable of converting complex queries into sub-queries. Below is a sub-table with rows randomly sampled from the original table. Based on the sub-table, decompose the original query into 2-3 complete sub-queries that can solve the original query.

\{{In-context examples}\}

Sub-table: \{{Sub-table}\}
Query:\{{Query}\}
New query: 
\end{lstlisting}
\end{tcolorbox}
\vspace{-3.5mm}
\caption{The prompt template for the query augmentor}
\label{fig:prompt-query}
\end{figure*}

\begin{figure*}[h]
\begin{tcolorbox}[left=1mm,right=1mm,top=0.mm, bottom=0mm,colback=white]
\begin{lstlisting}
============================== /*** Schema info ***/ ==============================
Instruction: Given the following table, you will add schema type about the columns in the table.
Schema type includes:
- Numerical: consists of digits and numerical symbols like decimal points or signs.
- Char: whether column content is a phrase or description.
- Date: whether column content represents time or date.

\{{In-context examples}\}

Table: \{{Sub-table}\}

============================== /*** Semantic info ***/ ==============================
Instruction: Given the following table, you need to first summarize the contents of the table, then based on the summary, give a concluded description of each of the columns.

\{{In-context examples}\}

Table: \{{Sub-table}\}
============================== /*** Literal info ***/ ==============================
Instruction: Below is a subtable with rows sampled, you are required to infer the data distribution and format from the sample data. Refine commonalities in literal representations within each table column.

\{{In-context examples}\}

Sub-table: \{{Sub-table}\}
\end{lstlisting}
\end{tcolorbox}
\vspace{-3.5mm}
\caption{The prompt template for three types of augmentation in the table augmentor}
\label{fig:prompt-aug}
\end{figure*}

\begin{algorithm*}[b]
\caption{ALTER Workflow}
\label{alg: alter}
\begin{algorithmic}[1]
\Require{original table-question pair ($T, Q$).}
\Ensure{predicted answer to the question $\hat{A}$.}
\Function{ALTER}{$T, Q$}

\State {\color{blue} \# \texttt{Function table organizer (Taborg) defined} }
\Function{TabOrg}{$T, Q$}  
    \State {\color{blue} \# \texttt{Store table augmentation information in advance} }
\State $Aug = \texttt{TabAug}$(T)
    \State $(Aug_{c_1}, \dots, Aug_{c_{|C|}}, Aug_T) \leftarrow Aug$
    \State {\color{blue} \# \texttt{Sample row index $\hat{R}$ using embedding-based similarity} }
    \State $\hat{R} =  \texttt{RowSample} (T, Q)$
    \State $\hat{C} = \texttt{ColFilter}(T_{\hat{R}, :}, Q, Aug)$
    \State sql = \texttt{CallLLM}$(T_{\hat{R},\hat{C}}, Q, Aug)$
    \State \textbf{return} \texttt{Execute}$(sql)$
\EndFunction
    \State {\color{blue} \# \texttt{Generate sub-queries with the query augmentor} }
    \State $\hat{R} = \texttt{RowSample}(T, Q)$
    \State $\{Q_i\}_{i=1}^m=$ \texttt{QueryAug}$(T_{\hat{R},:}, Q)$
    \State {\color{blue} \# \texttt{Run sub-queries in parallel} }
    \For{$i$ in $1,\cdots, m$} 
    \State $T_{i}^{res}$ = \texttt{Taborg}$(T, Q_i)$
    \State {\color{blue} \# \texttt{Get effective response for the sub-query} }
    \State $A^{res}_i = \texttt{CallLLM}(T_{i}^{res})$
    \EndFor
    \State {\color{blue} \# \texttt{Get accessible sub-table in the primary workflow} }
    \State $T^{res}$ = \texttt{Taborg}$(T, Q)$
    \State {\color{blue} \# \texttt{Joint reasoner} }
    \State $\hat{A}$ = \texttt{CallLLM}$(T^{res}, A^{res}_{1:m})$
    \State \textbf{return} $\hat{A}$
\EndFunction
\end{algorithmic}
\end{algorithm*}

\end{document}